\documentclass[conference]{IEEEtran}
\pdfoutput=1
\usepackage{cite}
\usepackage{amsmath,amsfonts,amsthm}
\usepackage{graphicx}
\usepackage{subcaption}
\usepackage[export]{adjustbox}
\usepackage{float}
\usepackage{pbox}
\usepackage{hyperref}
\usepackage{listings}
\usepackage{pifont}
\usepackage{xcolor}
\usepackage{courier}
\usepackage{soul}
\graphicspath{{figures/}}

\begin{document}
\title{A review of landmark articles in the field of co-evolutionary computing}

\author{\IEEEauthorblockN{Noe Casas}
\IEEEauthorblockA{ Email: research@noecasas.com}}

\maketitle


\begin{abstract}
Coevolution is a powerful tool in evolutionary computing that
mitigates some of its endemic problems, namely stagnation in local optima
and lack of convergence in high dimensionality problems.
Since its inception in 1990, there are multiple articles that have contributed
greatly to the development and improvement of the coevolutionary
techniques. In this report we review some of those landmark articles
dwelving in the techniques they propose and how they fit to
conform robust evolutionary algorithms.

\end{abstract}

\section{Introduction}

Evolutionary algorithms mimic some of the dynamics of natural evolution. This technical
report focuses on certain special forms of such dynamics: co-evolution. It is
defined as an evolutionary scenario where the survival and the very evolutionary changes
experiences by an individuals across it genetic path are affected by other individuals.\\

There is evidence of co-evolution in natural species. For instance, bees and flowers
mutually provide benefits: bees feed from the flower nectar while helping their
cross pollination. This is an example of cooperative coevolution. On the other hand,
species can also coevolve competitively, much like predators and preys, respectively
improving their attack and defense abilities. For instance, fossil records have
proven that snails shells have become thicker at the same rate the claws of their
predators have become stronger (and hence able to crush more easily the snail shell).\\

In order to study the topic, we have chosen seven articles that we consider to
be relevant to the development of the field's body of knowledge. The criteria used
to do the selection combine historical relevance with number of citations
(taken from google scholar\footnote{\url{https://scholar.google.es/}} and
citeseerx\footnote{\url{http://citeseerx.ist.psu.edu/}}).\\

This report is structured in three parts: first, section \ref{sec:review} provides
a thorough review of each of the selected articles, in chronological order;
then, section \ref{sec:discussion}
discusses their strong and weak points; finally, section \ref{sec:conclusions}
provide a brief overview of the state of the field, based on the reviewed
articles.

\section{Review of the selected articles} \label{sec:review}

\subsection{Co-evolving Parasites Improve Simulated Evolution As an
Optimization Procedure (Hillis, 1990)} \label{sec:hillis}

Hillis was the first to propose the use of co-evolution applied to evolutionary
computation in \cite{hillis90parasites} in 1990. His approach consisted of
a competitive evolution scenario where there are two species, referred to
as hosts-parasites or prey-predators, and where the goal of the system was
to improve the performance of \textit{sorting networks}.\\

\begin{figure}[H]
\centering
\def\svgwidth{.4\textwidth}
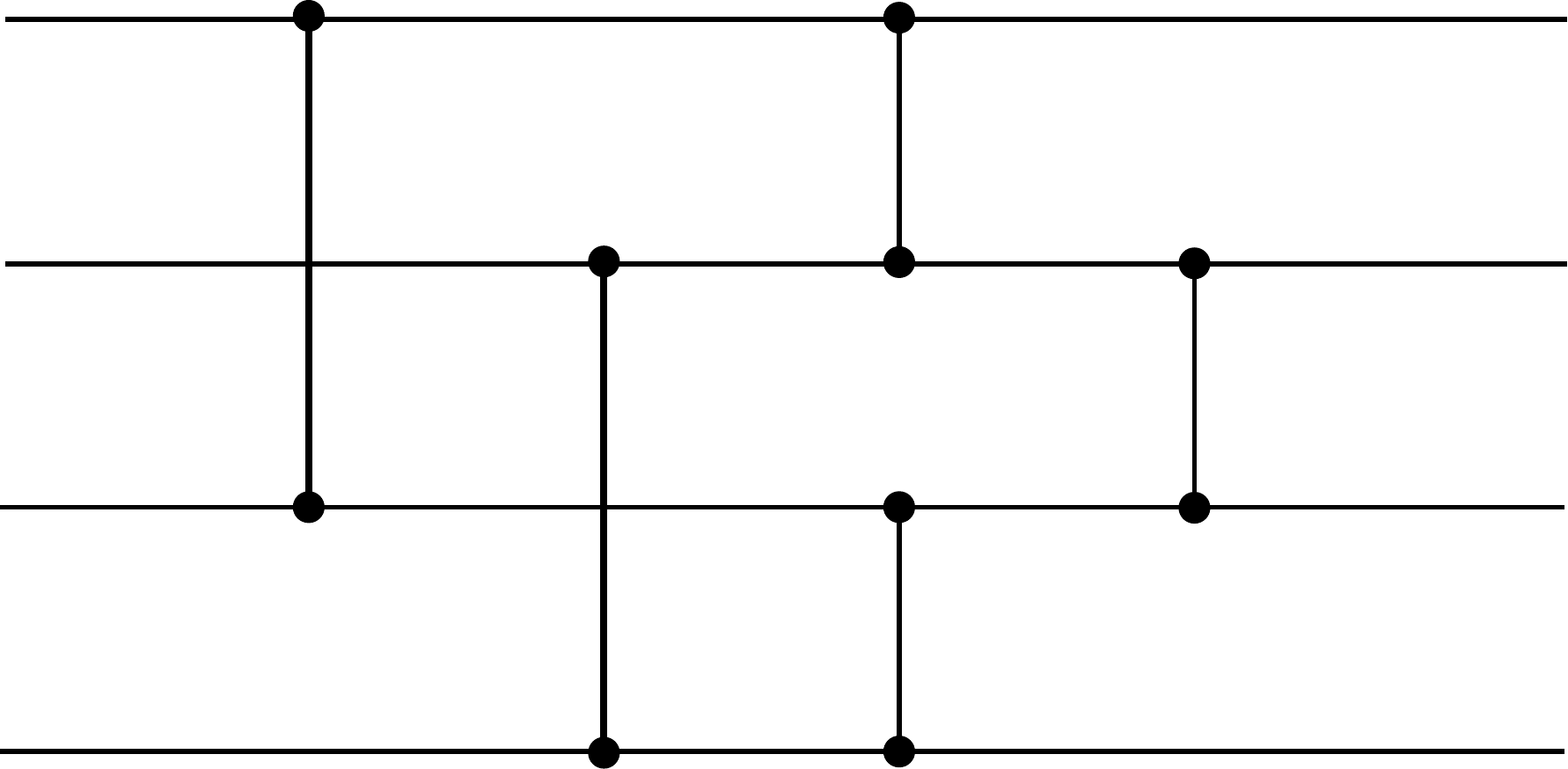
\caption{Structure of a 4-input sorting network.}
\label{fig:sortingnetwork}
\end{figure}

A sorting network is an algorithm that sorts the input data using fixed comparisons. They
differ from generic sorting algorithms in that they are not able to handle variable
number of input data and that their comparisons are pre-defined. They consist
of an assemblement of wires and comparators, as shown in figure \ref{fig:sortingnetwork}.\\

\begin{figure}[H]
\centering
\def\svgwidth{.4\textwidth}
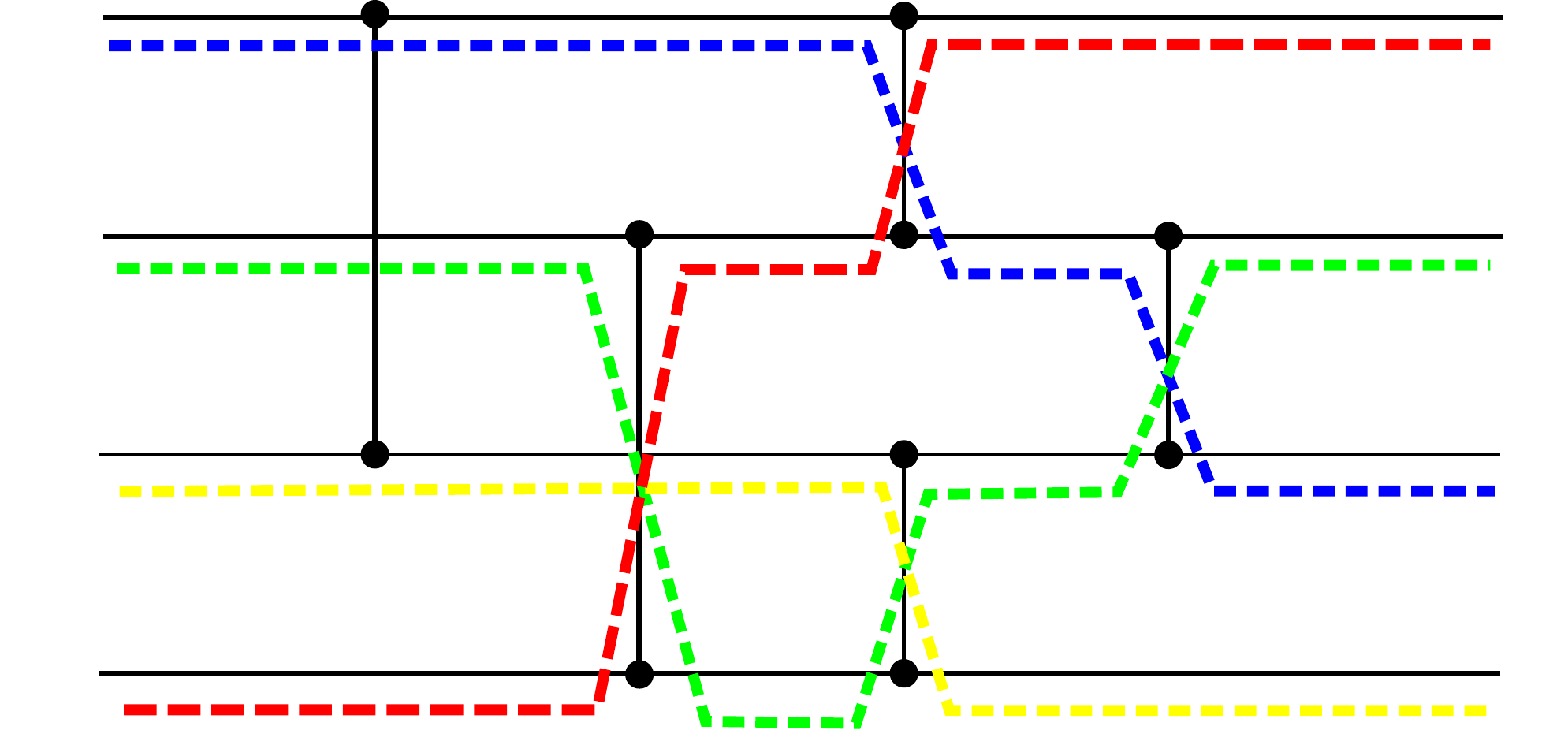
\caption{Behaviour of a sorting network.}
\label{fig:sortingnetworkoperation}
\end{figure}

The horizontal wires have the inputs of the netowrk on the left side and its outputs
(i.e. the sorted inputs) on the right. The vertical connectors are comparators
that \textit{switch} the wire data if the upper connection is greater than the other
one. This behaviour is illustrated in figure \ref{fig:sortingnetworkoperation}
\footnote{Images from mediawiki under license Creative Commons Attibution 3.0 Unported}.

The \textit{goodness} of a sorting network is judged based on its correction
(i.e. the results are always properly sorted) but also on the number of
comparators (i.e. the less comparators, the better), in order to minimize
execution time and cost (in case of deploying a hardware version of the
network).\\

It is worth mentioning that sorting networks have recently regained
popularity because they are used in GPU computing \cite{owens08gpu}.\\

On his first attempts to approach the problem without co-evolution,
Hillis found some of the typical problems attributed to soft
computing techniques: stagnation in local optima and overfitting.
In order to address these problems, Hillis introduced some changes to improve
population diversity, but only after introducing competitive co-evolution
he achieved notable results.\\

In the co-evolutionary algorithm proposed by Hillis, \textit{hosts}
represented configurations of 
the sorting network, while \textit{parasites} represented test data
to be supplied to a host as input. The fitness of each group is opposed
to the other group: the fitness of the hosts depends on how many test cases
(i.e. parasites) an individual has succeeded in sorting, while the fitness
of the parasites depends on how many times it made a host fail sorting.\\

Despite the validity and novelty of the approach proposed by Hillis, we believe
his article deserve some fair criticism:

\begin{itemize}

\item From the literary composition point of view, Hillis' paper lacks proper structure. It
has three sections named "Introduction", "Sorting networks" and "The co-evolution of
parasites". In order to better deliver the information it contains, it should
contain a section devoted to explain the experiments, another section exposing their
results and another one drawing its conclusions and describing future research lines.\\

\item From the scientific point of view, Hillis paper lacks proper presentation of the
results and proper measurement of their performance. One would expect tables
and graphics with comparisons of different alforithm tunings and argumentations for
misbehaviours of the algorithms. All the appreciations in the article do not seem
to be the result of rigorous study but appear rather \textit{intuitive} and loose.

\end{itemize}

\subsection{Co-evolutionary constraint satisfaction (Paredis, 1994)}

Jan Paredis applied in \cite{paredis94constraint} the approach pioneered by Hillis
to a Constraint Satisfaction Problem (CSP).\\

The motivation for proposing a coevoling approach is that 
constraint satisfaction
GA techniquest used in previous studies 
(i.e. Genetic Repair \cite{muhlenbein92parallelgenetic},
Decoders that always lead to valid representations \cite{davis85epistatic}, 
Penalty Functions \cite{richardson89guidelines}) where all problem specific.\\

This way, he devised a new generic approach for constraint handling termed
\textit{Co-evolutionary Constraint Satisfaction} (CCS). In this approach,
there is no domain knowledge to actively enforce the satisfaction of constraints,
but only checks to know whether they are met.\\

The problem we tried to solve is the N queens arrangement on a chess board. Paredis'
algorithm maintains two populations: one with potential solutions and another one
with constraints. They act much like Hillis' predator-prey model, leading to an
"arms race" between both populations.\\

Although the presentation and structure of the article are correct, the
presentation of its results lacks proper measurements of the algorithm performance.

\subsection{Evolving Complex Structures via Cooperative Coevolution
(De Jong and Potter, 1995)} \label{sec:dejongpotter}

De Jong and Potter used coevolution in \cite{jong95evolvingcomplex} to address the
problem of designing evolutionary solutions that exhibit modularity (as oposed to
the classic evolutionary algorithm where the solution \textit{emerges as-is}).\\

Their generic approach consists in maintaining different species evolving separately in
their own populations (following a schema inspired in the \textit{island model}
\cite{grosso85island}). Each of these species represent a submodule that can be
combined with the others to form a solution to the problem. Apart from them, there
is an extra species that merges representative individuals from
the former populations into a single individual, which is then subject to
evalutation. Its credit flowed back to the original component individuals' fitness.\\

The intent of this approach is to provide the individual species selective
pressure to cooperate instead of compete, while keeping the competition among
individuals of the same population.\\

They applied this generic schema to two different problems: function optimization
and robot task learning.\\

Regarding function optimization, De Jong and Potter have each species to determine
a specific parameter of the function to be optimized. When combined, they conform
the complete parameter set that fully qualifies the function solution. They
tested their approach using two functions: the highly multimodal Rastrigin function
and the Rosenblock function \footnote{The Rosenblock function was originally defined
in the De Jong test function suite \cite{dejong75analysis}.}, defined by two
highly interdependent variables with a very deep valley containing the global optimum.
Their results were mixed, as some of the variations of their approach outperformed
a standard GA while others did not.\\

Regarding the robot learning problem, they used a system that maintained a model
of the world and a set of production rules that were fed into a rules engine
in order to determine the proper actions to be performed. In this case the
genetic algorithm is in charge of evoling the rule set, taking into account
the feedback loop to evaluate the fitness of the newly evolved rules. The different
populations created different rule sets that were later combined to conform
the robot's behaviour.\\

The structure of the article is the standard one, with different sections to
describe the problems to be solved, the results obtained, and a discussion
about them. De Jong and Potter use best fitness value plots to compare the performance
of their architecture with other reference approaches. However, they fail to
provide evidence of the benefits of their proposal, but only sketch its
aparent potential, leaving more concrete results for further studies.\\

Potter and De Jong continued this research line, both exploring individually
the experiments performed in this article, like in \cite{potter94function}
and \cite{potter95acoevolutionary}, and dwelving deeper
into the generic sub-component evolutionary framework and extending its
applicability to other domains, like in \cite{potter00cooperativecoevolution}.

\subsection{New Methods for Competitive Coevolution (Rosin and Belew, 1996)}
\label{sec:rosinbelew}

In their article \cite{rosin96newmethods}, Rosin and Belew used
competitive coevolution in the frame of game theory, following the
previous experiences of John Maynard Smith \cite{smith82games} and
Robert Axelrod \cite{axelrod87evolution}.\\

The games \textit{Nim} and \textit{3D tic-tac-toe} are used as problems
to test the techniquest used in this article. The authors use the term
\textit{host} to refer to individuals whose fitness it being tested,
and the term \textit{parasite} for individuals used to the the hosts'
fitness.\\

Three novel co-evolutionary techniquest were introduced by the authors
in this article:

\begin{itemize}
\item \textit{Competitive Fitness Sharing}: defeating a parasite awards an amount
of points, that are shared among all hosts that defeated it. This rewards
hosts that can defeat parasites that no other host could beat, hence
decreasing the probability of parasites that no host can defeat.
\item \textit{Shared Sampling}: in order to keep the needed computational power
as low as possible, not all hosts would fight against every parasite,
but only against a sample of them. In order to select a strong parasite
sample, the individual that defeated most opponents during the previous
generation is selected, let us call it A; then, those hosts beating
parasites that defeated A are selected until the sample is large enough.
\item \textit{Hall of Fame}: given the fact that we use finite populations but
we do not want to lose strong parasites, a record is kept with the best
parasite in each generation. Hosts are tested against the current parasite
generation and the aforementioned best parasites of all times, which are
referred to as \textit{hall of fame}. They play a role analogous to that
of elitism, but with the purpose of improving the testing instead of
improving the population fitness.
\end{itemize}

The article provides a thorough study of the dynamics of the populations
under the aforementioned techniques, identifying equilibrium conditions
as well as extinction probabilities.\\

We consider this paper by Rosin and Belew to be of utmost quality,
both regarding its correct structuration and exposition of their ideas,
and also on the scientific evaluation and argumentation of the different
techniquest and the comparison of their performances.

\subsection{Coevolving Predator and Prey Robots: Do "Arms Races" Arise
in Artificial Evolution? (Nolfi and Floreano, 1998)} \label{sec:nolfifloreano}
%

Stefano Nolfi and Dario Floreano researched in \cite{nolfi98coevolving} the so-called
"arms race" in competitive coevolutionary populations. This term refers to the
property of those populations for trying to get fitter over generations
to beat the individuals from the other populations; given that 
populations' fitnesses are coupled by the competition, an increase in the fitness
of one population usually leads to a decrease in the fitness of the other
populations, normally alternating such a trend cyclically.\\

For their experiments, the authors used the classical Kephera robots
to set up a predator-prey scenario. The intelligence of the robot
is implemented in a perceptron with recurrent connections in the output
layer. In order to determine the network weights, Nolfi and Floreano
use an evolutionary algorithm where such weights are encoded as
alleles in the genome (i.e. direct encoding of the neural network).
The role of the predator is to detect the prey with its
sensors and chase it until touching it. Two populations are
maintained: one to optimize the weights of the
predator and another one for the prey, both co-evolving in competition,
using simple fitness functions: for the predator, 1 if it catches
the prey, 0 otherwise; vice versa for the prey.\\

The authors first define the metric under which they shall
evaluate their techniques. For this, they
study the effects of the \textit{Red Queen Effect},
by which the evolutionary benefits obtained by some individuals are
reduced or eliminated by other population; this is possible because
the fitness of an individual is also coupled to its competitors'
performance. To avoid this problem, they propose a variation to
the \textit{Current Individual vs. Ancestral Opponents} \cite{cliff95tracking},
called \textit{Master Tournament}, which consists of
testing the performance of the best individual of each
generation against each best competitor of all generations.\\

They then elaborate on the proposal of Rosin and Belew \cite{rosin96newmethods}
(covered in section \ref{sec:rosinbelew}), hipothesizing that their
\textit{Hall of fame} approach may progressively lead to having less
and less selective pressure to devise strategies effective against
the current enemies and more and more biased towards past generations.\\

Their conclusions are that 
continuous increase in objective \textit{goodness} is not guaranteed
by competitive co-evolution, as
populations can cycle between strategies that only provide temporary
advantage over the other populations and not long-term improvements.
That this effect can be reduced by keeping ancestors to test
individuals from the current generation, but this may hinder
the effects of co-evolution themselves, because it leads to give more
and more importance to devising a strategy that is successful with
the ancestors instead of the current generation.

Although the article contains plenty of information and data, the
organization makes it less than obvious to fit together the conclusions
from each section. Furthermore, despite they state that the goal
of the article is to determine the credibility of the "arms race"
hipotheses, other topics are mixed in the exposition of the experiments
(e.g. \textit{Red Queen Effect}, \textit{Hall of Fame} problems,
the \textit{Bootstap} problem) without proper
introduction or justification. We believe that the contents of the paper
would have benefited remarkably from a classical section arrangement
that would have enabled clearer conveyance of the authors' conclusions.

%
%
%
%
%

\subsection{Pareto coevolution: Using performance against coevolved opponents
in a game as dimensions for Pareto selection (Noble and Watson, 2001)}

Jason Noble and Richard A. Watsonstudied in \cite{noble01paretocoevolution}
the applicability of Pareto optimality to coevolutionary algorithms. They
use this idea to play Texas Hold'em Poker. Each individual encodes in
its genome the strategies to play under different conditions, and
plays against its fellow players, getting fitness reward in case of
winning.\\\

Their approach to introduce Pareto Optimality in a coevolutionary
scenario is to consider each oponent to be a dimension that has to be
optimized. Given that a Pareto-optimal solution is one where none
of the dimensions can be improved without damaging the performane the
other dimensions, trying to find the pareto front actually means
devising individuals that \textit{play well} against all other oponents.\\

In order to avoid the \textit{red queen effect} (see section
\ref{sec:nolfifloreano}), the add to the set of opponents some
\textit{reference poker strategies}, that is, hardcoded (by themselves)
poker strategies that remained constant over time. They were
not optimal in any case, but they acted as
a \textit{fixed reference point} to objectively measure the fitness
of the population.\\

Although their approach overperformed a normal GA, their results were
not remarkably good, but they introducing Pareto optimality
in the coevolutionary frame proved to be a successful line of
research (e.g. \cite{ficici01pareto}).
%
%
%
%
%

\section{Discussion} \label{sec:discussion}

As seem in the summary of the selected articles presented in section
\ref{sec:review}, there are two different branches for coevolutionary
algorithms: competitive and cooperative.\\

In \textbf{competitive coevolution} approaches there are several subpopulations
having the fitness of the individuals of each population defined so
that advantage of one population implies disadvantage of the others.
Under certain conditions, this leads to an "arms race" among the
subpopulations, which is presumed to mitigate the stagnation in local
minima -typical from classical evolutionary algorithms- thanks to
the variation in time of the overall fitness landscape. The different
subpopulations normally play different roles, like predator-prey
(e.g. \cite{nolfi98coevolving} in section \ref{sec:nolfifloreano})
or host-parasite (e.g. \cite{hillis90parasites} in section
\ref{sec:hillis}). Among such roles, we tend to see one that represents
the individuals being tested (i.e. predator, host) and one that 
represents the tests themselves (i.e. prey, parasite). This way,
improving in the fitness of a test means that the tested individual
performs worse, and vice versa.\\

Although the arms race usually leads to improvement of the fitness,
there are some \textbf{pathological behaviours} in this approach. The first
one is the situation where the populations cycle over alternative
fitness increasing periods, but do not achieve \textit{objective
improvement}; this problem is usually referred to as
\textit{mediocre objective stasis}. This, as pointed out by
Floreano and Nolfi in \cite{nolfi98coevolving}, can be mitigated
by keeping an archive of ancestor opponents (e.g. \textit{Hall of fame})
and by introducing some objetive measurements that evaluate the
objective improvement of the populations, like the reference
poker strategies by Noble and Watson in \cite{noble01paretocoevolution}.
These two techniques also help with another typical problem
of competitive coevolution: the lose of gradient, commonly
referred to as the \textit{Red Queen Effect}, which happens when
one population achieves a level so superior compared to the other,
that nothing can be learned by either population by competing.\\
Another problem in competitive coevolution is the \textit{lack of promotion}
of strong tests (i.e. tests that make tested individuals perform poorly).
This happens either because tested individuals are not paired against
them or because their effect in the fitness of tested individuals
is very small due to the larger amount of favorable tests.
For these two problems, Rosin and Belew in \cite{rosin96newmethods}
used respectively \textit{Competitive Fitness Sharing} and
\textit{Shared Sampling}. The former makes that the easier a test is
(e.g. a lot of testing individuals performed well against it),
the less fitness it awards. The latter makes that tests are
paired with the individuals that have shown weaker against them
in previous generations. Multiobjective evolution has also been
used in this regard
(e.g. \cite{noble01paretocoevolution}) by having each test
be considered one of the criteria to be optimized. Hence, finding
the Pareto front is analogous to find oponents that perform
well against all tests.\\

In \textbf{cooperative coevolution} approaches there are several
populations that combine their efforts to achieve fitter
solutions. Their most typical use case is where each population
has individuals of certain species that only represent part
of the final solution. Representatives of each species are
then combined to form a complete solution and the credit
derived from its fitnes flows down to the original subparts.
This effectively consists in decomposing the problem in
modular parts, hence achieving reduction in the dimensionality
of the original solution space.\\

\section{Conclusions} \label{sec:conclusions}

We have traversed some of the remarkable articles regarding
coevolutionary algorithms. Competitive and cooperative
approaches offer complementary techniques to address
complex problems where conventional evolutionary computation
falls short regarding stagnation in local optima.\\

Despite this progress, there is still a significant need
for crafting the algorithm and tuning it with problem-dependent
considerations.

\bibliographystyle{IEEEtran}
\bibliography{biblio}

\end{document}